\pdfoutput=1

\documentclass[11pt]{article}

\usepackage[preprint]{acl}

\usepackage[utf8]{inputenc} 
\usepackage[T1]{fontenc}    
\usepackage{hyperref}       
\usepackage{url}            
\usepackage{booktabs}       
\usepackage{amsfonts}       
\usepackage{nicefrac}       
\usepackage{microtype}      
\usepackage{xcolor}         
\usepackage{graphicx}
\usepackage{multirow} 
\usepackage{subcaption} 
\usepackage{listings}
\usepackage{longtable}
\usepackage{tcolorbox}
\usepackage{colortbl}
\usepackage{changepage}
\usepackage{amsmath}
\usepackage{pifont}

\title{Theorem-of-Thought: A Multi-Agent Framework for Abductive, Deductive, and Inductive Reasoning in Language Models}

\author{Samir Abdaljalil\textsuperscript{*}, Hasan Kurban\textsuperscript{†}, Khalid Qaraqe\textsuperscript{†}, Erchin Serpedin\textsuperscript{*} \\
  \textsuperscript{*}Texas A\&M University, College Station, TX., USA\\
  \textsuperscript{†}Hamad Bin Khalifa University, Doha, Qatar\\
  \texttt{sabdaljalil@tamu.edu, hkurban@hbku.edu.qa} \\}

\begin{document}
\maketitle

\begin{center}
   {\large \textbf{Abstract}}
\end{center}

\begin{adjustwidth}{0.5cm}{0.5cm}
Large language models (LLMs) have shown strong performance across natural language reasoning tasks, yet their reasoning processes remain brittle and difficult to interpret. Prompting techniques like Chain-of-Thought (CoT) enhance reliability by eliciting intermediate reasoning steps or aggregating multiple outputs. However, they lack mechanisms for enforcing logical structure and assessing internal coherence. We introduce \textbf{Theorem-of-Thought (ToTh)}, a novel framework that models reasoning as collaboration among three parallel agents, each simulating a distinct mode of inference: abductive, deductive, and inductive. Each agent produces a reasoning trace, which is structured into a formal reasoning graph. To evaluate consistency, we apply Bayesian belief propagation guided by natural language inference (NLI), assigning confidence scores to each step. The most coherent graph is selected to derive the final answer. Experiments on symbolic (\textsc{WebOfLies}) and numerical (\textsc{MultiArith}) reasoning benchmarks show that ToTh consistently outperforms CoT, Self-Consistency, and CoT-Decoding across multiple LLMs, while producing interpretable and logically grounded reasoning chains. Our findings suggest a promising direction for building more robust and cognitively inspired LLM reasoning. 
The implementation is available at \url{https://github.com/KurbanIntelligenceLab/theorem-of-thought}.

\end{adjustwidth}

\section{Introduction}
\label{sec:intro}

Large language models (LLMs) have achieved impressive performance across a wide range of natural language understanding and generation tasks~\citep{wang-etal-2024-rethinking-bounds}, enabled by advances in in-context learning~\citep{sia2024incontext}, instruction tuning~\citep{zhang2024instructiontuninglargelanguage}, and chain-of-thought (CoT) prompting~\citep{cot}. These methods have extended LLMs' capabilities to handle complex forms of reasoning, including mathematical, logical, and commonsense inference.

Despite these advances, LLM reasoning remains shallow and unreliable. Existing approaches often rely on single-shot or sampling-based decoding along linear reasoning paths, making them susceptible to hallucinations~\citep{abdaljalil2025sindexsemanticinconsistencyindex}, logical inconsistencies~\citep{uceda2024reasoning}, and weak generalization~\citep{liu-etal-2025-zero}. Methods such as CoT and Self-Consistency~\citep{cot, wang-consistency} encourage intermediate steps and majority voting across sampled outputs, but lack mechanisms to verify internal coherence and model the logical structure of reasoning. As a result, outputs may appear fluent and plausible while remaining logically unsound.

This brittleness contrasts sharply with human reasoning, which is inherently multifaceted. Drawing on insights from cognitive science~\citep{okoli_reasoning}, we observe that human inference typically blends three complementary modes—abduction, deduction, and induction—that support explanation, derivation, and generalization. However, LLMs typically conflate these distinct processes into a single, undifferentiated flow, limiting both interpretability and reliability.

To address this gap, we propose \textbf{Theorem-of-Thought (ToTh)}, a framework that models diverse reasoning strategies through structured, verifiable interactions. ToTh employs three specialized agents, each emulating a distinct cognitive mode:
\begin{itemize}
    \item \textbf{Abduction}: inferring plausible explanations for observed facts;
    \item \textbf{Deduction}: deriving valid conclusions from given premises;
    \item \textbf{Induction}: generalizing from patterns or examples.
\end{itemize}

Each agent independently generates a reasoning trace, which is transformed into a \textbf{Formal Reasoning Graph (FRG)}—a directed graph where nodes represent intermediate conclusions and edges capture logical dependencies. We evaluate the internal consistency of each FRG using Bayesian belief propagation, with edge confidence scores calibrated via a Natural Language Inference (NLI) model. A composite score balancing average belief and logical entropy is used to select the most coherent graph, from which the final answer is extracted.

\paragraph{Contributions.} The key results of this work are:
\begin{itemize}
    \item We introduce ToTh, a structured reasoning framework that integrates abductive, deductive, and inductive inference into a modular LLM-based pipeline.
    \item We develop a belief propagation mechanism over reasoning graphs, leveraging NLI to assess and score logical coherence through Bayesian updates.
    \item We demonstrate that ToTh consistently outperforms state-of-the-art reasoning methods (e.g., CoT, Self-Consistency, CoT-Decoding) across multiple LLMs.
    \item Our evaluation on symbolic (\textsc{WebOfLies}) and numerical (\textsc{MultiArith}) benchmarks highlights ToTh’s robustness on tasks requiring multi-step inference—settings where direct prompting often fails~\citep{allen-zhu2025physics}.
\end{itemize}

The remainder of the paper is organized as follows: Section~\ref{sec:related} reviews related work. Section~\ref{sec:method} presents the ToTh framework. Section~\ref{sec:exp} describes the experimental setup, and Section~\ref{sec:results} analyzes the results obtained.  Section~\ref{sec:conclusion} concludes with implications for structured reasoning in LLMs and future research directions.

\section{Related Work}
\label{sec:related}

\paragraph{Prompt-based Reasoning in LLMs.} 
A growing body of work explores prompting strategies to enhance the reasoning capabilities of LLMs. CoT prompting~\citep{cot} encourages models to decompose problems into intermediate steps, guiding reasoning along a linear path. Building on this, Auto-CoT~\citep{zhang2023autocot} automates prompt generation by sampling diverse questions and producing corresponding reasoning traces, reducing manual effort. Beyond prompt generation, several works focus on optimizing prompt selection strategies. ActivePrompt~\citep{diao-etal-2024-active} identifies high-uncertainty instances for annotation, improving data efficiency and reasoning robustness through active learning. More recent approaches introduce explicit structure into the reasoning process. Tree-of-Thought (ToT)~\citep{tot23} enables multi-path exploration with internal evaluation, while Graph-of-Thought (GoT)~\citep{yao-etal-2024-got} structures reasoning as a graph to better model dependencies between steps.

\paragraph{Instruction Tuning for Reasoning.} 
Instruction tuning and knowledge distillation offer alternative approaches to eliciting reasoning in LLMs without relying on explicit prompts~\citep{lobo-etal-2025-impact, ranaldi-freitas-2024-self, lai-nissim-2024-mcot}. While effective, these methods typically require computationally intensive fine-tuning on large-scale datasets annotated with reasoning traces and CoT examples, which are often costly and domain-specific. Recent work has explored more indirect supervision strategies. For instance, \citet{liu2024proxytuning} introduce proxy tuning, which leverages auxiliary models to contrast a base LLM with its adapted variant. Although this approach reduces the need for direct supervision, it still assumes access to CoT-like outputs and pre-aligned reasoning benchmarks.



\begin{figure*}[ht]
\centering
\includegraphics[width=\textwidth]{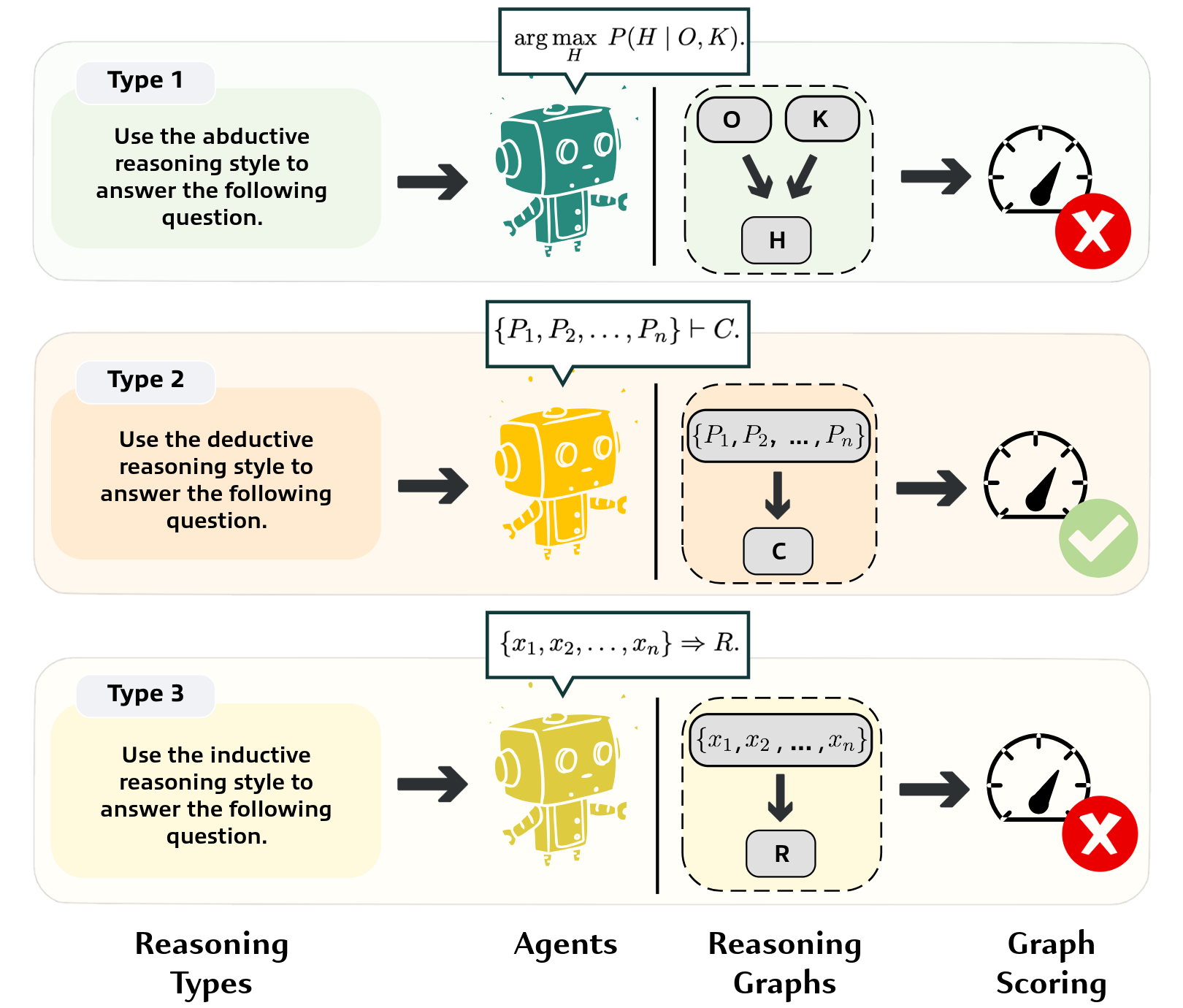} 
\caption{Overview of the Theorem-of-Thought (ToTh) reasoning pipeline. A question is independently processed by three agents, each using a distinct reasoning style: abductive (Type 1), deductive (Type 2), and inductive (Type 3). Each agent produces a structured reasoning graph, which is scored via Bayesian confidence propagation. Abduction infers the best hypothesis \( H \) given observations \( O \) and knowledge \( K \) (i.e., \( \arg\max_H P(H \mid O, K) \)); deduction derives a conclusion \( C \) from premises \( \{P_1, \dots, P_n\} \) (i.e., \( \{P_i\} \vdash C \)); induction generalizes from examples \( \{x_1, \dots, x_n\} \) to a rule \( R \) (i.e., \( \{x_i\} \Rightarrow R \)). The highest-scoring graph contributes its final node as the answer.  \ding{51} and \ding{55} indicate whether a given agent’s output was selected.
}
\label{fig:method}
\end{figure*}


\section{Methodology}
\label{sec:method}

ToTh is a graph-based reasoning framework designed to enhance the accuracy, interpretability, and generalization capabilities of LLMs on complex tasks. It decomposes reasoning into three modular agents, each simulating a classical inference paradigm—abduction, deduction, and induction. Each agent produces a structured reasoning trace, which is composed into a FRG. Final answers are derived via NLI-calibrated Bayesian belief propagation and composite graph scoring. The full pipeline is depicted in Fig.~\ref{fig:method}.

ToTh differs from prior reasoning paradigms along three axes: architecture, supervision, and verification. Prompt-based methods (e.g., CoT, ToT, GoT) elicit reasoning via linear or loosely structured traces, yet lack mechanisms for enforcing logical consistency. Instruction-tuned models embed reasoning behavior through fine-tuning on annotated traces, often requiring large datasets and remaining opaque at inference time. While both families reflect growing interest in structured multi-step reasoning, they typically operate within monolithic or implicit architectures and do not support formal consistency checking. In contrast, ToTh instantiates distinct cognitive agents, integrates their outputs into an interpretable graph, and explicitly verifies reasoning coherence through NLI-guided Bayesian inference—enabling modular, transparent, and verifiable reasoning beyond the scope of existing methods.

\paragraph{Multi-Paradigm Reasoning Agents.} Given a natural language question \( q \), ToTh deploys three independent solver agents, each aligned with a distinct classical mode of inference: abductive, deductive, and inductive reasoning. These paradigms are formally defined as follows.

The abductive reasoning agent \( a_1 \) infers the most plausible hypothesis \( H \) given a set of observations \( O \) and background knowledge \( K \), formalized as:
\[
a_1: \quad \arg\max_H \; P(H \mid O, K).
\]
The deductive reasoning agent \( a_2 \) derives a conclusion \( C \) that logically follows from a set of premises \( \{P_1, P_2, \dots, P_n\} \), represented as:
\[
a_2: \quad \{P_1, P_2, \dots, P_n\} \vdash C.
\]
The inductive reasoning agent \( a_3 \) generalizes a rule \( R \) from observed examples \( \{x_1, x_2, \dots, x_n\} \), expressed as:
\[
a_3: \quad \{x_1, x_2, \dots, x_n\} \Rightarrow R.
\]
Each agent \( a_i \in \{a_1, a_2, a_3\} \) independently produces a reasoning trace 
\[
\mathbf{r}^{(i)} = \left[r^{(i)}_1, r^{(i)}_2, \dots, r^{(i)}_{s_i} \right],
\]
where \( r^{(i)}_j \) denotes the \( j \)-th step in the agent’s reasoning process. 

\paragraph{Formal Reasoning Graph Construction.} Each reasoning trace \( \mathbf{r}^{(i)} \) is transformed into a directed graph \( G^{(i)} = (V^{(i)}, E^{(i)}) \), where \( V^{(i)} \) denotes the set of nodes representing individual reasoning steps, and \( E^{(i)} \) represents directed edges encoding inferential relationships between those steps.  Edges \( (v_u \rightarrow v_v) \in E^{(i)} \) are inferred using a pretrained NLI model, which assesses the semantic relationship between reasoning steps. Each edge is annotated with a trust score \( \theta_{uv} \in [0, 1] \) based on the predicted label:

\[
\theta_{uv} =
\begin{cases}
0.95 & \text{if entailment} \\
0.60 & \text{if neutral} \\
0.10 & \text{if contradiction}
\end{cases}
\]

These scores quantify the strength of logical entailment between intermediate steps, providing a calibrated basis for probabilistic reasoning in the subsequent belief propagation stage.

\paragraph{Bayesian Confidence Propagation.} 
To model belief flow across the graph, belief values are propagated using a Bayesian update rule, adapted from classical formulations of belief propagation in probabilistic graphical models~\citep{pearl1988probabilistic}.

Each node \( v \in V \) is initialized with a prior confidence \( P(v) = 0.5 \), reflecting maximum uncertainty. For a node \( v_c \) with a single parent \( v_p \) and associated trust score \( \theta_{pc} \), the updated belief is computed using a Bayesian update rule:

\[
P(v_c) = \frac{P(v_p) \cdot \theta_{pc}}{P(v_p) \cdot \theta_{pc} + (1 - P(v_p)) \cdot (1 - \theta_{pc})}.
\]

In the case of multiple parents \( \{v_{p_1}, \dots, v_{p_m}\} \), the belief for \( v_c \) is computed as the average of individual updates from each parent:

\[
P(v_c) = \frac{1}{m} \sum_{j=1}^{m} \, f\big(P(v_{p_j}), \theta_{p_jc}\big)
\]
\[
f(p, \theta) = \frac{p \cdot \theta}{p \cdot \theta + (1 - p)(1 - \theta)} \; .
\]

This recursive formulation propagates confidence through the graph, amplifying agreement across consistent reasoning paths while attenuating belief when upstream uncertainty or contradiction is detected.

\paragraph{Graph Scoring.} Each reasoning graph \( G^{(i)} \) is evaluated based on a trade-off between average node confidence and logical uncertainty. We prioritize graphs that are both confident (high belief) and low in uncertainty (low entropy). The mean confidence is computed as
\[
\mu^{(i)} = \frac{1}{|V^{(i)}|} \sum_{v \in V^{(i)}} P(v),
\]
and the normalized binary entropy is given by
\[
H^{(i)} = -\frac{1}{|V^{(i)}|} \sum_{v \in V^{(i)}} h(P(v))
\]
\[
h(p) = p \log p + (1 - p) \log (1 - p) \; . 
\]
The final score combines both terms:
\[
\text{Score}(G^{(i)}) = \mu^{(i)} - H^{(i)}.
\]
The reasoning graph with the highest score is selected as the final candidate:
\[
G^* = \arg\max_i \, \text{Score}(G^{(i)}).
\]

\paragraph{Answer Extraction.} The final answer is extracted from the terminal node of the selected graph \( G^* \), corresponding to the last step in the associated reasoning trace.

\paragraph{Theoretical Complexity.}
\label{sec:complexity}

Let \( k = 3 \) denote the number of reasoning agents, and \( s \) the number of reasoning steps generated per agent. The ToTh framework involves three main stages of computation: trust estimation, belief propagation, and graph scoring. During trust estimation, each agent produces a sequence of reasoning steps, and an NLI model is applied to each adjacent pair to evaluate the strength of logical connection. Since each trace contains at most \( s - 1 \) such pairs, the total number of NLI evaluations across all agents is \( \mathcal{O}(k \cdot s) \). In the belief propagation stage, each node in the constructed reasoning graphs is visited exactly once in topological order, and its posterior confidence is updated based on incoming trust scores using a Bayesian update rule, resulting in \( \mathcal{O}(k \cdot s) \) total updates. Finally, graph scoring involves computing the average confidence and entropy over all nodes in each graph, which also requires \( \mathcal{O}(k \cdot s) \) time. Therefore, the end-to-end complexity of the ToTh pipeline is \( \mathcal{O}(k \cdot s) \), linear in both the number of agents and the number of reasoning steps per agent.

This makes ToTh substantially more efficient than sampling-based methods such as Self-Consistency or CoT-Decoding, which require \( \mathcal{O}(n) \) decoding passes, where \( n \) is the number of sampled reasoning chains. In contrast, ToTh executes a single, structured reasoning pass per agent, followed by lightweight verification and scoring, offering a more scalable and interpretable alternative to stochastic decoding.

\section{Experiments}
\label{sec:exp}

\paragraph{Data.} 
ToTh was evaluated on two representative reasoning benchmarks. \textsc{MultiArith}~\citep{roy-etal-2015-reasoning} targets compositional numerical inference through multi-step arithmetic word problems. \textsc{WebOfLies}~\citep{suzgun-etal-2023-challenging}, part of the \textsc{BIG-Bench-Hard} suite, involves determining truth values among logically entangled symbolic statements. These datasets are known to challenge LLMs under direct prompting~\citep{allen-zhu2025physics}, making them suitable for testing structured reasoning capabilities.

\begin{figure*}[ht]
\centering
\includegraphics[width=1\textwidth]{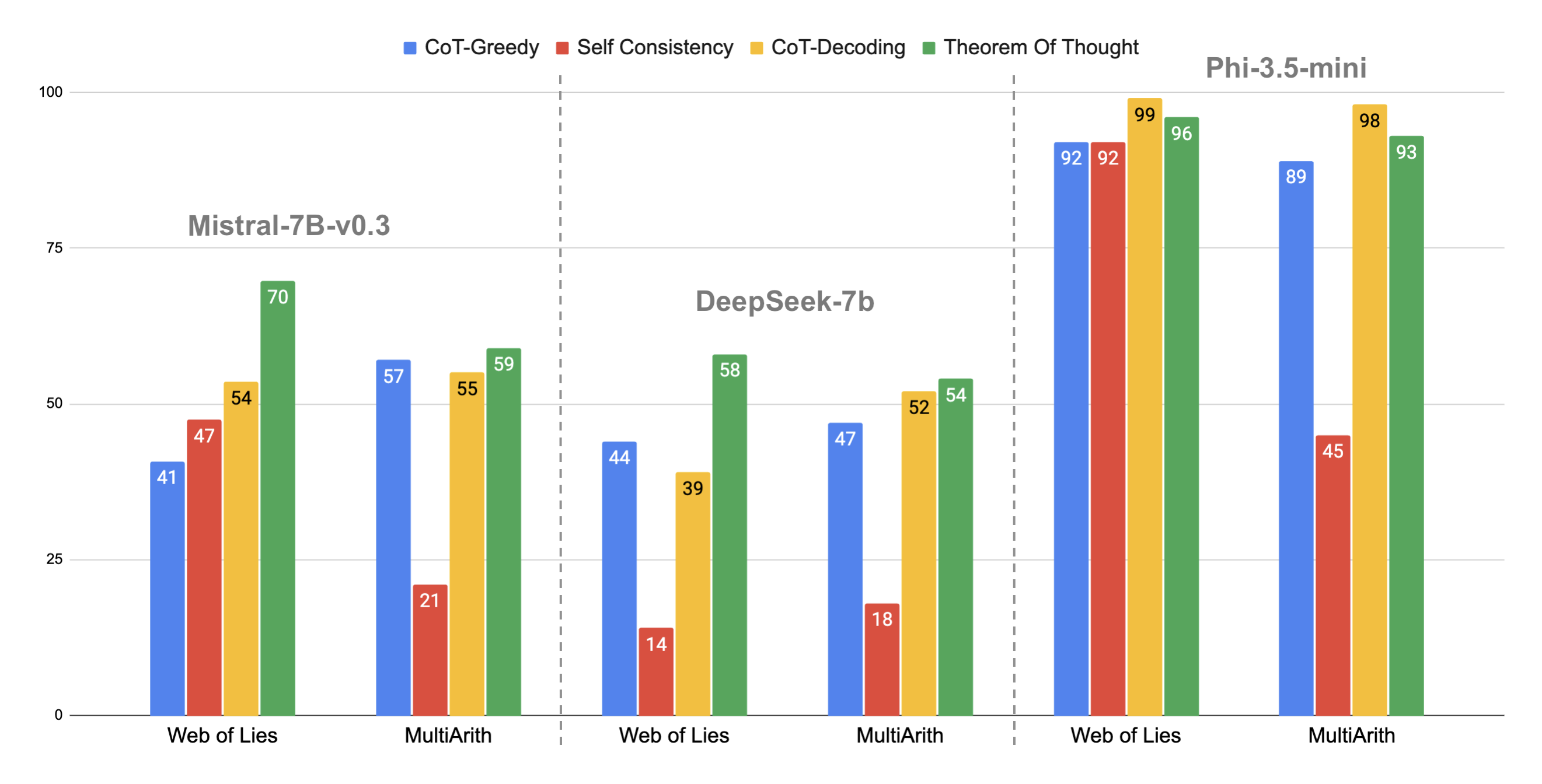} 
\caption{Accuracy (\%) comparison across reasoning pipelines on two benchmark tasks (\textsc{WebOfLies} and \textsc{MultiArith}) using three open-source language models: \textsc{Mistral-7B-v0.3}, \textsc{DeepSeek-7B}, and Phi-3.5-mini. Each group of bars corresponds to a different reasoning method: CoT-Greedy (blue), Self-Consistency (red), CoT-Decoding (yellow), and our proposed Theorem-of-Thought (green).}
\label{fig:results}
\end{figure*}

\paragraph{Models.} 
Three publicly available LLMs were selected to provide diversity in scale, alignment, and architecture: (1) \textsc{Mistral-7B}~\citep{jiang2023mistral7b}\footnote{\url{https://huggingface.co/mistralai/Mistral-7B-Instruct-v0.3}}, a general-purpose decoder model with efficient scaling; (2) \textsc{DeepSeek-7B}~\citep{deepseekai2025deepseekr1incentivizingreasoningcapability}\footnote{\url{https://huggingface.co/deepseek-ai/deepseek-llm-7b-chat}}, an instruction-tuned model optimized for multi-turn reasoning and alignment; and (3) \textsc{Phi-3.5 Mini}~\citep{abdin2024phi3technicalreporthighly}\footnote{\url{https://huggingface.co/microsoft/Phi-3.5-mini-instruct}}, a lightweight model designed for educational, low-cost reasoning tasks. This selection spans compact inference-efficient models to instruction-aligned reasoning-focused systems.

\paragraph{Baselines.} 
ToTh was compared with three strong baselines: CoT~\citep{cot}, Self-Consistency~\citep{wang-consistency}, and CoT-Decoding~\citep{wang2024cotdecoding}. CoT prompts the model to generate intermediate reasoning steps before answering. Self-Consistency improves robustness by sampling \( n = 20 \) completions and selecting the most frequent answer. CoT-Decoding eliminates explicit prompting by using diverse decoding paths to stimulate latent reasoning behaviors.

\paragraph{Experimental Setup.} 
All models were evaluated in their released form without fine-tuning. Decoding was performed with temperature 0.7 and a maximum output length of 526 tokens. RoBERTa-MNLI\footnote{\url{https://huggingface.co/FacebookAI/roberta-large-mnli}} was used for scoring reasoning coherence, consistent with prior work on NLI-based output validation~\citep{farquhar2024detecting}. Inputs were uniformly formatted as ``Q: [question] \textbackslash n A:'' across all methods for consistency with baselines~\citep{wang2024cotdecoding}.

To direct reasoning behavior, the following instruction was prepended to each input, with the appropriate \texttt{\{style\}} keyword for each agent:

\begin{quote}
\texttt{Use the \textbf{\{style\}} reasoning style to answer the following question.} \\
\texttt{Follow these instructions carefully:}
\begin{itemize}
    \item \texttt{Break the problem into clear, numbered reasoning steps using \{style\}.}
    \item \texttt{Reference any known principles, patterns, or assumptions involved.}
    \item \texttt{Arrive at a final answer that directly responds to the question.}
\end{itemize}
\end{quote}

All experiments used a single decoding pass per input. Random seeds were fixed, and decoding settings were held constant for reproducibility.

\section{Results}
\label{sec:results}

\subsection{Main Experimental Results}
Results are reported as answer accuracy (\%) and summarized in Figure~\ref{fig:results}.
\paragraph{Performance Across Models.} 
ToTh consistently outperforms all baseline methods on both tasks when evaluated with \textsc{Mistral-7B} and \textsc{DeepSeek-7B}, demonstrating clear gains in reasoning accuracy. On \textsc{Phi-3.5 Mini}, although CoT-Decoding marginally surpasses ToTh on certain instances, ToTh maintains consistently strong performance across both symbolic and numerical tasks. For example, on the \textsc{WebOfLies} dataset, ToTh improves over CoT-Greedy by 29\% and 14\% on \textsc{Mistral-7B} and \textsc{DeepSeek-7B}, respectively, and remains within 3\% of the top-performing method on \textsc{Phi-3.5 Mini}. These results highlight ToTh’s robustness and generalization across models of varying scale and alignment.

\paragraph{Comparison with CoT-Decoding.} While CoT-Decoding performs strongly on Phi-3.5-mini, achieving near-perfect scores on \textsc{WebOfLies} (99\%), ToTh achieves comparable or slightly lower performance (96\%) while maintaining higher consistency across models. For example, on the \textsc{MultiArith} dataset, ToTh surpasses CoT-Decoding by 4–5 points on both \textsc{Mistral-7B} and \textsc{DeepSeek-7B}, indicating superior generalization in numerical reasoning.

\paragraph{Self-Consistency Under-performance.} Surprisingly, Self-Consistency under-performs across all settings, particularly on symbolic tasks. For instance, it yields only 14\% and 21\% on \textsc{WebOfLies} and \textsc{MultiArith} with \textsc{DeepSeek-7B} and \textsc{Mistral-7B}, respectively. This suggests that majority-vote over stochastic generations fails to capture structured dependencies, especially in logic-heavy tasks.

\paragraph{Model Sensitivity.} As expected, performance scales with model capability. Phi-3.5-mini achieves the highest absolute scores across all methods, reflecting its stronger alignment and training. However, ToTh’s margin over baselines remains meaningful even at lower model scales, suggesting that the architecture contributes to reasoning robustness beyond just model size. While \textsc{DeepSeek-7B} is trained with reasoning capabilities in mind, its broader training objectives, including code generation and open-ended question answering, may diffuse its specialization in structured reasoning tasks. In contrast, Phi-3.5-mini benefits from a targeted curriculum focused on educational and step-by-step problem-solving, which likely accounts for its superior performance on both symbolic and mathematical benchmarks. Interestingly, \textsc{Mistral-7B} consistently outperforms \textsc{DeepSeek-7B} despite being similar in size. This may be attributed to Mistral’s cleaner, reasoning-focused pretraining data and architecture-level optimizations, which enhance its ability to follow multi-step instructions and maintain logical coherence across token spans.

\subsection{Robustness Under Reasoning Complexity}

To evaluate the robustness of ToTh under increasing reasoning complexity, experiments were conducted using the \textsc{Mistral-7B} model on both symbolic and numerical tasks. Table~\ref{tab:difficulty_levels} presents accuracy results stratified by problem difficulty: the number of interdependent statements (3–5) for \textsc{WebOfLies}, and operation depth/length combinations for \textsc{MultiArith}.

ToTh maintains strong performance across all difficulty levels, outperforming or closely matching leading baselines. In symbolic reasoning, ToTh achieves 43\% accuracy on the most challenging setting (5 statements), significantly exceeding CoT-Greedy (19\%) and Self-Consistency (38\%), and closely approaching CoT-Decoding (46\%). This trend persists across simpler instances, where ToTh attains the highest scores at 3 and 4 statements.

For numerical reasoning, ToTh delivers the strongest results at lower complexity levels—achieving state-of-the-art performance at $d_0$/$l_3$ (59\%) and $d_0$/$l_4$ (45\%)—and remains competitive even at higher complexity ($d_2$/$l_3$), with accuracy comparable to CoT-Decoding (21\% vs. 24\%). These findings highlight ToTh’s capacity to generalize across task difficulty and suggest that its structured, multi-agent reasoning design offers a scalable advantage under increased inference load.

\begin{table}[t]
\small
\centering
\setlength{\tabcolsep}{3pt}
\renewcommand{\arraystretch}{1.1}
\begin{tabular}{lcccccc}
\toprule
 & \multicolumn{3}{c}{\textsc{WebOfLies}} & \multicolumn{3}{c}{\textsc{MultiArith}} \\
  & 3 & 4 & 5 & $d_0/l_3$ & $d_0/l_4$ & $d_2/l_3$ \\
\cmidrule(lr){2-4} \cmidrule(lr){5-7}
CoT-G   & 41 & 32 & 19 & 57 & 26 & 14 \\
SelfC   & 48 & 47 & 38 & 21 & 6 & 17 \\
CoT-Dec & 54 & 48 & \textbf{46} & 55 & 41 & \textbf{24} \\
ToTh    & \textbf{70} & \textbf{56} & \underline{43} & \textbf{59} & \textbf{45} & \underline{21} \\
\bottomrule
\end{tabular}
\vspace{0.5em}
\caption{\small Accuracy (\%) of \textsc{Mistral-7B} on symbolic (\textsc{WebOfLies}) and mathematical (\textsc{MultiArith}) reasoning tasks across increasing levels of difficulty. Columns 3–5 correspond to symbolic reasoning with 3, 4, and 5 interdependent statements, respectively. Columns $d_0/l_3$, $d_0/l_4$, and $d_2/l_3$ represent arithmetic reasoning problems categorized by depth and length: $d$ denotes operation depth and $l$ indicates sequence length. ToTh achieves the highest accuracy in 5 out of 6 settings and remains competitive even on the most complex instances, demonstrating consistent performance across symbolic and numerical domains. \textbf{Bold}: best performance; \underline{Underlined}: second-best.}

\label{tab:difficulty_levels}
\end{table}

\section{Conclusion and Future Work}
\label{sec:conclusion}

This work presents Theorem-of-Thought (ToTh), a graph-based reasoning framework that integrates abductive, deductive, and inductive inference through a modular multi-agent design. Each agent generates structured reasoning traces, which are composed into formal graphs and verified using NLI-calibrated Bayesian confidence propagation. This approach supports both accurate prediction and interpretable, logically grounded reasoning. Empirical evaluations on symbolic and numerical benchmarks demonstrate that ToTh consistently outperforms strong prompting and decoding baselines, particularly in scenarios requiring structured logical inference.

ToTh introduces a new paradigm in reasoning with language models by treating inference as a verifiable, compositional process, rather than a monolithic generation task. Future research will explore dynamic agent routing based on input characteristics, inter-agent collaboration protocols, and adaptive trust estimation via fine-tuned and ensemble-based NLI models. Extending the framework to scientific hypothesis validation, law and policy reasoning, and multimodal domains such as visual question answering represents a promising direction for advancing general-purpose, verifiable reasoning in large language models.

\section*{Limitations}
\label{sec:limitations}

\paragraph{Fixed Reasoning Types.}
ToTh presumes a uniform decomposition into abductive, deductive, and inductive reasoning across all inputs. While this modularity improves interpretability, it imposes a fixed cognitive scaffold that may not align with tasks requiring hybrid or atypical inference patterns. For example, creative tasks or ambiguous prompts may benefit from dynamically blending reasoning types or emphasizing one over others. This rigidity can limit ToTh’s adaptability and lead to suboptimal trace composition in such cases. Future work may explore data-driven and  context-sensitive agent routing, allowing the framework to selectively instantiate and suppress reasoning paradigms based on input semantics.

\paragraph{Propagation Sensitivity.}
The Bayesian confidence propagation mechanism is sensitive to noise in low-confidence nodes, which may attenuate otherwise valid reasoning chains or distort belief estimates in deeper regions of the graph. This can occur in longer traces where errors in early reasoning steps propagate disproportionately, reducing the reliability of final predictions. Moreover, current propagation is uniform and unregularized, lacking robustness mechanisms against adversarial and  inconsistent intermediate steps. Incorporating calibrated uncertainty modeling, edge dropout, and confidence smoothing—potentially informed by fine-grained entailment distributions—could enhance stability and mitigate the amplification of localized inconsistencies.

\bibliographystyle{acl_natbib}




\end{document}